\documentclass[conference]{IEEEtran}
\DeclareUnicodeCharacter{03BC}{\textmu}

\IEEEoverridecommandlockouts
\usepackage{amsmath,amssymb,amsfonts}
\usepackage{algorithmic}
\usepackage{graphicx}
\usepackage{textcomp}
\usepackage{xcolor}

\usepackage{hyperref}
\usepackage[acronym]{glossaries}
\newacronym{FSM}{FSM}{Finite State Machine}
\newacronym{SNN}{SNN}{Spiking Neural Network}
\newacronym{CU}{CU}{Control Unit}
\newacronym{DP}{DP}{Datapath}
\newacronym{LUT}{LUT}{Look-Up Table}
\newacronym{AER}{AER}{Address Event Representation}
\newacronym{SAIF}{SAIF}{Switching Activity Interchange Format}
\newacronym{LIF}{LIF}{Leaky Integrate and Fire}
\newacronym{FPGA}{FPGA}{Field Programmable Gate Array}
\newacronym{NN}{NN}{Neural Network}
\newacronym{FF}{FF}{Flip Flop}
\newacronym{wrt}{w.r.t.}{with respect to}
\newacronym{DSP}{DSP}{Digital Signal Processor}
\newacronym{BRAM}{BRAM}{Block RAM}
\newcommand{\mycomment}[1]{}
\usepackage[style=ieee, maxbibnames=1, minbibnames=1]{biblatex}

\def\BibTeX{{\rm B\kern-.05em{\sc i\kern-.025em b}\kern-.08em
    T\kern-.1667em\lower.7ex\hbox{E}\kern-.125emX}}
\begin{document}
\glsdisablehyper

\title{Energy-Efficient Digital Design: A Comparative Study of Event-Driven and Clock-Driven Spiking Neurons
\thanks{This work was partially supported by project SERICS (PE00000014) under the MUR National Recovery and Resilience Plan funded by the European Union.}
\thanks{To foster research in this field, we are making our experimental code available as open source: \url{https://github.com/smilies-polito/spiking-neurons-comparison.git} }
}

\author{\IEEEauthorblockN{Filippo Marostica, Alessio Carpegna, Alessandro Savino, Stefano Di Carlo}
\IEEEauthorblockA{Control and Computer Engineering Department, Politecnico di Torino, Turin, Italy\\
Email: \{filippo.marostica, alessio.carpegna, allesandro.savino, stefano.dicarlo\}@polito.com}
}

\maketitle

%% in the reference capisci come fare per citare solo il primo autore - cerca online
%% si possono togliere anche il link o avaialbe at link ...
%% compattare tabelle in una unica

\begin{abstract}
% 118 words - limit is 200
\glsresetall
This paper presents a comprehensive evaluation of \gls{SNN} neuron models for hardware acceleration by comparing event-driven and clock-driven implementations. We begin our investigation in software, rapidly prototyping and testing various \gls{SNN} models—based on different variants of the \gls{LIF} neuron—across multiple datasets. This phase enables controlled performance assessment and informs design refinement. Our subsequent hardware phase, implemented on \gls{FPGA}, validates the simulation findings and offers practical insights into design trade-offs. In particular, we examine how variations in input stimuli influence key performance metrics such as latency, power consumption, energy efficiency, and resource utilization. These results yield valuable guidelines for constructing energy-efficient, real-time neuromorphic systems. Overall, our work bridges software simulation and hardware realization, advancing the development of next-generation \gls{SNN} accelerators.
\glsresetall
\end{abstract}


\section{Introduction}

%% How do hardware-inspired approximations in event-driven neuron models affect the performance of SNN implementations, and under which conditions do these methods offer tangible benefits over traditional clock-driven approaches?

In recent years, \glspl{SNN} have emerged as a promising paradigm for energy-efficient computation, particularly when implemented using dedicated neuromorphic hardware \cite{carpegna_spiker_2024}. Unlike conventional neural networks that use continuous activation functions, \glspl{SNN} emulate the discrete, event-based operation of biological neurons by transmitting information through spikes. %\cite{shahsavari_advancements_2023}. 
This temporal coding mechanism enables \glspl{SNN} to exploit the inherent sparsity of neural activity, potentially reducing energy consumption and improving processing speed for time-dependent tasks.

Within the realm of digital \glspl{SNN} hardware accelerators, two primary processing methods have been explored: clock-driven and event-driven \cite{isik_survey_2023}. The clock-driven approach updates neuron states at each clock tick, % regardless of whether they receive inputs or not,
conversely, the event-driven method updates neurons only when an input spike occurs reducing unnecessary computations and better mimics the behavior of biological neurons \cite{brette2007simulation}.

%In this context, our work tries to better understand how to identify the optimal processing method for \gls{SNN} neurons. Selecting the most suitable model depends on balancing trade-offs among data characteristics, hardware constraints, and performance metrics. To address this, we performed an extensive cross-layer analysis. We resorted to complete \gls{SNN} software models to evaluate the accuracy of different neuron models under varied conditions. In parallel, we implemented these neuron designs in hardware on \glspl{FPGA} to precisely measure metrics such as latency, power, energy consumption, and resource utilization. By comparing the event-driven and clock-driven implementations, we analyzed how predefined architectural parameters—such as processing mode (event vs. clock driven) and input sparsity—affect neuron behavior, providing practical insights for designing energy-efficient neuromorphic systems.

This work explores the trade-offs between event-driven and clock-driven processing for \gls{SNN} neurons through a cross-layer analysis. We evaluate accuracy using software models and assess latency, power, energy, and resource usage via hardware implementations on \glspl{FPGA}. By varying key architectural parameters—such as processing mode and input sparsity—we provide practical insights into the design of energy-efficient neuromorphic systems.

The paper is organized as follows: Section~\ref{sec:neuron_model} discusses the neuron model and its event-driven and clock-driven implementations, Section~\ref{sec:software_implementation} focuses on the software implementation and accuracy analysis of the neuron models, Section~\ref{sec:hardware_implementation} covers the hardware implementation on FPGA, Section~\ref{sec:results} presents the results and performance analysis, finally, Section~\ref{sec:future_works} concludes the paper and outlines potential future work.

\section{Neuron model}\label{sec:neuron_model}
%% STRUCTURE
%   - Two levels description of the analysys --> focus on lower level (neuron)
%   - General description of different neuron models
%   - Finish with the LIF neuron

Recent research in neuromorphic computing emphasizes efficient hardware platforms \cite{pavanello_special_2023}, exploring optimization across multiple abstraction levels \cite{yamazaki_spiking_2022}. High-level optimizations focus on the network architecture and parameters \cite{electronics13091744}, while lower-level optimization addresses fundamental computational elements like neuron models and connections. Our work focuses on the latter, analyzing how neuron processing methods impact performance.
Neuron models have historically balanced biological accuracy and computational efficiency. Early biophysically detailed models like Hodgkin-Huxley \cite{hodgkin_quantitative_1952} were computationally demanding, prompting simplified alternatives. Among these, the \gls{LIF} model \cite{brette_adaptive_2005} captures essential neuronal dynamics efficiently, making it suitable for neuromorphic hardware. The next sections present the mathematical formulation of the standard and quantized \gls{LIF} models for hardware acceleration.

\subsection{Mathematical Model of the LIF Neuron}\label{sec:math_model_neuron}
%% STRUCTURE
%   - Mathematical formulation of the model

The \gls{LIF} model is derived from an equivalent electrical circuit where the neuron’s membrane is represented as a capacitor $C$ in parallel with a resistor $R$, modeling ion leakage. The membrane potential $U_{\text{mem}}(t)$ follows the differential equation: \[ \tau \frac{dU_{\text{mem}}(t)}{dt} = - U_{\text{mem}}(t) + RI_{\text{in}}(t) \quad (where\; \tau= RC) \] The neuron exhibits a threshold-based firing mechanism: when $U_{\text{mem}}(t)$ exceeds a predefined threshold $U_{\text{threshold}}$, the neuron emits a spike and resets its membrane potential.

\subsection{Event Driven and Clock Driven implementation}\label{subsec:ED_and_CD_implementation}

For digital hardware implementations, the continuous-time model solution is discretized, obtaining: \[ {U_{\text{mem}}[t + \alpha \cdot t] = \beta U_{\text{mem}}[t] + W X[t] - R[t]} \]
Where $W$ is the weight, $X[t]$ the input at time $t$, $R[t]$ the reset applied on spike, and $\beta = e^{-\frac{\alpha \cdot t}{\tau}}$ is the exponential decay coefficient.

Both clock-driven and event-driven models compute the exponential decay of the membrane potential using the decay factor $\beta$, but they differ in how and when this evaluation occurs.
In the clock-driven approach, the neuron updates its membrane potential at every clock cycle by multiplying it by $\beta$. In the literature, this has been implemented either with an exact multiplier or with a shifter that approximates the multiplication by the nearest power of two \cite{carpegna_spiker_2022,carpegna_spiker_2024}.
In contrast, the event-driven approach updates the membrane potential only when an event occurs. This requires evaluating the time elapsed between two consecutive spikes ($\Delta t = t_{i+1} - t_i$). In the absence of input spikes, the membrane potential follows an exponential decay governed by the recursive relation: $U_{\text{mem}}(t+1) = \beta U_{\text{mem}}(t)$. By iterating over two time steps where no input occurs, this extends to: $U_{\text{mem}}(t+2) = \beta^2 U_{\text{mem}}(t)$; and generalizing on $n$ time steps between two active input spikes: $U_{\text{mem}}(t+n) = \beta^n U_{\text{mem}}(t)$.

Since \( \Delta t \) is variable, the computation of \( \beta^{\Delta t} \) must be efficient. To achieve this, a \gls{LUT} is typically used, storing precomputed values of \( \beta^{\Delta t} \) for different intervals. As in the clock-driven case, two implementations are possible: (1) storing exact values of \( \beta^{\Delta t} \), functionally equivalent to using a multiplier, and (2) approximating \( \beta^{\Delta t} \) with the closest power of two, enabling a hardware-efficient shifter-based implementation.

%Since \( \Delta t \) is variable, \( \beta^{\Delta t} \)is typically retrieved from a \gls{LUT} with precomputed values. As in the clock-driven case, this can be implemented either with exact values (multiplier-based) or approximated by powers of two for efficient shifting.

%Besides mathematical differences, event-driven neurons differ from clock-driven implementations regarding how inputs are processed. Clock-driven neurons handle inputs in a serial manner, while event-driven ones handle inputs either serially or through \gls{AER}, a protocol widely used in neuromorphic systems to encode spikes as addressed events. Although \gls{AER} can be adapted to clock-driven systems, its event-driven nature makes it more efficient for event-driven designs; the overhead in clock-driven implementations would render it less practical for our purposes. Indeed, \gls{AER} directly interfaces with neuromorphic sensors like silicon retinas \cite{lichtsteiner_128times_2008} or silicon cochleae \cite{liu_event-based_2010}, significantly reducing bandwidth by avoiding continuous data streams. Based on these options, to comprehensively evaluate these design choices, our comparative analysis was conducted using six different configurations: (i) clock-driven serial neuron with an exact multiplier, (ii) clock-driven serial neuron with a shifter, (iii) event-driven serial neuron with an exact multiplier, (iv) event-driven serial neuron with a shifter, (v) event-driven \gls{AER} neuron with an exact multiplier, and (vi) event-driven \gls{AER} neuron with a shifter.

Besides their mathematical differences, event-driven neurons differ from clock-driven ones in how inputs are handled. Clock-driven neurons process inputs serially, while event-driven neurons can do so either serially or via \gls{AER}, a protocol widely used in neuromorphic systems that encodes spikes as addressed events. Although \gls{AER} can be adapted to clock-driven systems, its overhead makes it more efficient for event-driven designs. \gls{AER} also directly interfaces with neuromorphic sensors (e.g., silicon retinas \cite{lichtsteiner_128times_2008} or silicon cochleae \cite{liu_event-based_2010}), reducing bandwidth by avoiding continuous data streams. 
Our comparative analysis examined six configurations: (i) clock-driven serial neuron with an exact multiplier, (ii) clock-driven serial neuron with a shifter, (iii) event-driven serial neuron with an exact multiplier, (iv) event-driven serial neuron with a shifter, (v) event-driven \gls{AER} neuron with an exact multiplier, and (vi) event-driven \gls{AER} neuron with a shifter.

\section{Software Implementation}\label{sec:software_implementation}

The first phase of this work focuses on analyzing the impact of the considered neuron models on the accuracy of \glspl{SNN} built on top of them. While software models cannot fully replicate the exact behavior and constraints of hardware, they are valuable tools for validating the computational model, analyzing the accuracy in specific tasks (e.g., classification), and identifying potential bottlenecks.
%In this context, software models are valuable tools for validating the computational model, analyzing the accuracy in specific tasks, and identifying potential bottlenecks.

%\subsection{Structure of the Custom Neuron Model}

For this software-level analysis, the SnnTorch \cite{eshraghian_training_2023} Python framework was selected due to its seamless integration with PyTorch \cite{paszke_pytorch_2019}. Along with the default SnnTorch's \gls{LIF} neuron a custom event-driven neuron was developed by overriding the standard \gls{LIF} one. For this preliminary software-level analysis, the difference in spike encoding between serial and \gls{AER} is irrelevant and, therefore, not considered.

%Figure~\ref{fig:sw_flow} shows the two main components of this custom neuron: (i) a customizable \gls{LIF} neuron based on SnnTorch, and (ii) a mask generation process. The mask, computed during input pre-processing, flags nonzero inputs at each time step ($M[t]=1$ if any input is nonzero; otherwise, $M[t]=0$). Event-driven neurons process inputs only when $M[t]=1$; otherwise, they increment a counter without computation, emulating spike-driven behavior. The clock-driven model uses the same mask to ensure identical stimulus patterns, facilitating an accurate comparison of computational behavior and accuracy between both approaches.

\begin{figure}[t]
\centerline{\includegraphics[width=\columnwidth]{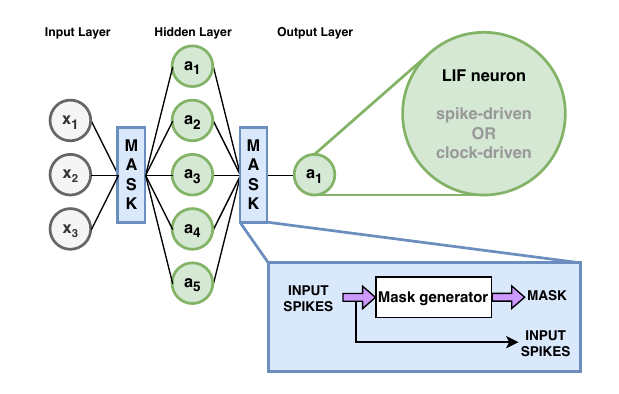}}
\caption{High-level view of an SNN architecture incorporating a custom mask layer to track input activity an two custom neuron layers configurable to operate either in a spike-driven or clock-driven mode.}
\label{fig:sw_flow}
\end{figure}

Figure~\ref{fig:sw_flow} illustrates the custom neuron model, consisting of a configurable \gls{LIF} neuron and a mask generator. The mask flags active inputs at each time step, allowing event-driven neurons to process only when needed, while ensuring consistent input patterns across both neuron types for fair comparison.

\subsection{Sparsity Analysis on datasets}

Our analysis compares three datasets reflecting different spiking-data scenarios (see Table~\ref{tab:accuracy_density}). The datasets are characterized by different input densities along two dimensions—temporal density (i.e., frequency of updates) and per-time step input activity (i.e., active inputs per timestep)—obtained by analyzing a subset of 1,000 elements from each dataset. These benchmarks cover three common scenarios—spiking by nature, static-to-spike conversion, and dynamic-to-spike conversion. Figure~\ref{fig:all_metrics} illustrates their differing temporal densities, with each dataset corresponding to a distinct band.
MNIST \cite{deng_mnist_2012} comprises static images converted to spikes via Poisson rate-coding, yielding nearly 100\% temporal density. N‑MNIST \cite{noauthor_frontiers_nodate} is inherently spiking data recorded by a dynamic ATIS sensor, and AudioMNIST \cite{becker_audiomnist_2023} includes non-spiking audio data requiring spike conversion. 

\subsection{Experimental Validation and Results}

%To assess the performance of the different neuron models, we started comparing their impact on inference accuracy on the three considered datasets while training the networks under identical conditions using the standard (clock-driven) neuron integrated in SnnTorch.
%
%This analysis is crucial not only for studying the effects of using approximated hardware models but also for determining the best approach to managing these methods within a full network. The idea of this study is to investigate how approximated decay coefficients impact accuracy in event-driven neuron implementations and to compare those findings with similar clock-driven neurons. Specifically, we focus on identifying and minimizing potential errors introduced by the shifter-based approximation of the decay coefficient.

To evaluate different neuron models, we compared their effects on inference accuracy across three datasets under identical training conditions with SnnTorch’s standard (clock-driven) neuron. This analysis clarifies how approximate hardware models behave and guides their integration into full networks. 
We specifically examine how approximated decay coefficients affect accuracy in event-driven neurons compared to clock-driven ones, focusing on minimizing errors introduced by shifter-based approximations.

\begin{table}[t]
\caption{Clock-driven vs. Event-driven Neurons Inference Accuracy and Dataset Characteristics}
\begin{center}
\begin{tabular}{|c|c|c|c|}
\hline
\textbf{} & \textbf{MNIST} & \textbf{N-MNIST} & \textbf{AudioMNIST} \\
\hline
\multicolumn{4}{|c|}{\textbf{Inference Accuracy (\%)}} \\
\hline
Clock-driven (Multiplier) & 99.22 & 96.24 & 96.09 \\
\hline
Clock-driven (Shifter) & 99.22 & 96.24 & 96.09 \\
\hline
Event-driven (Multiplier) & 99.22 & 96.24 & 96.09 \\
\hline
Event-driven (Shifter) & 99.22 & 96.23 & 96.09 \\
\hline
\multicolumn{4}{|c|}{\textbf{Dataset Properties (\%)}} \\
\hline
Temporal Density & $100 \pm 0$ & $93.7 \pm 3.80$ & $16.6 \pm 2.8$ \\
\hline
Input Density & $13.2 \pm 5.1$ & $1.6 \pm 0.5$ & $74.8 \pm 12.7$ \\
\hline
\end{tabular}
\label{tab:accuracy_density}
\end{center}
\end{table}

In multiplier-based neurons, any value of $\beta$ can be used without significant hardware or algorithmic constraints. However, in shifter-based neurons, an ``optimal" $\beta$ must be chosen to minimize accuracy loss. Typically, $\beta$ is near 1, and to reduce quantization errors when approximating with powers of 2, we define $\beta = 1 - \beta'$,  where $\beta'$ is a power of 2 (i.e., $\beta' = 2^{-n}$). For instance, $\beta = 0.9325$ ($\beta' = 0.0625 = 2^{-4}$) for the AudioMNIST and MNIST datasets, and $\beta = 0.5$ ($\beta' = 0.5 = 2^{-1}$) for N-MNIST. Although this approach increases area slightly (due to extra logic), it significantly reduces the error compared with a simpler shifter-based scheme.

%Both shifter-based and multiplier-based event-driven neurons rely on a \gls{LUT} to approximate $\beta^{\Delta t}$, introducing small inaccuracies depending on temporal sparsity and specific LUT entries. Input Sparsity and Temporal Sparsity notably influence neuron behavior. For dense datasets like MNIST, event-driven and clock-driven neurons show equivalent accuracy; more temporally sparse datasets (N-MNIST, AudioMNIST) reveal slight discrepancies due to their inherent sparsity and model sensitivities.

%Despite these theoretical inaccuracies, software simulations indicate minimal practical accuracy differences between event-driven and clock-driven neurons (Table~\ref{tab:accuracy_density}). The shifter-based approximation offers a practical trade-off by reducing quantization error with a modest area overhead. 

Despite minor inaccuracies from \gls{LUT}-based approximations of $\beta^{\Delta t}$, software simulations show minimal accuracy differences between event-driven and clock-driven neurons. While sparsity affects behavior slightly—especially in sparse datasets—the shifter-based design provides a good balance between efficiency and accuracy. Future work could examine network training directly on quantized models to further improve precision. Overall, shifter-based neurons perform comparably to multiplier-based versions across various tasks and data sets.

\section{Hardware Implementation}\label{sec:hardware_implementation}

\begin{figure}[t]
\centerline{\includegraphics[width=\columnwidth]{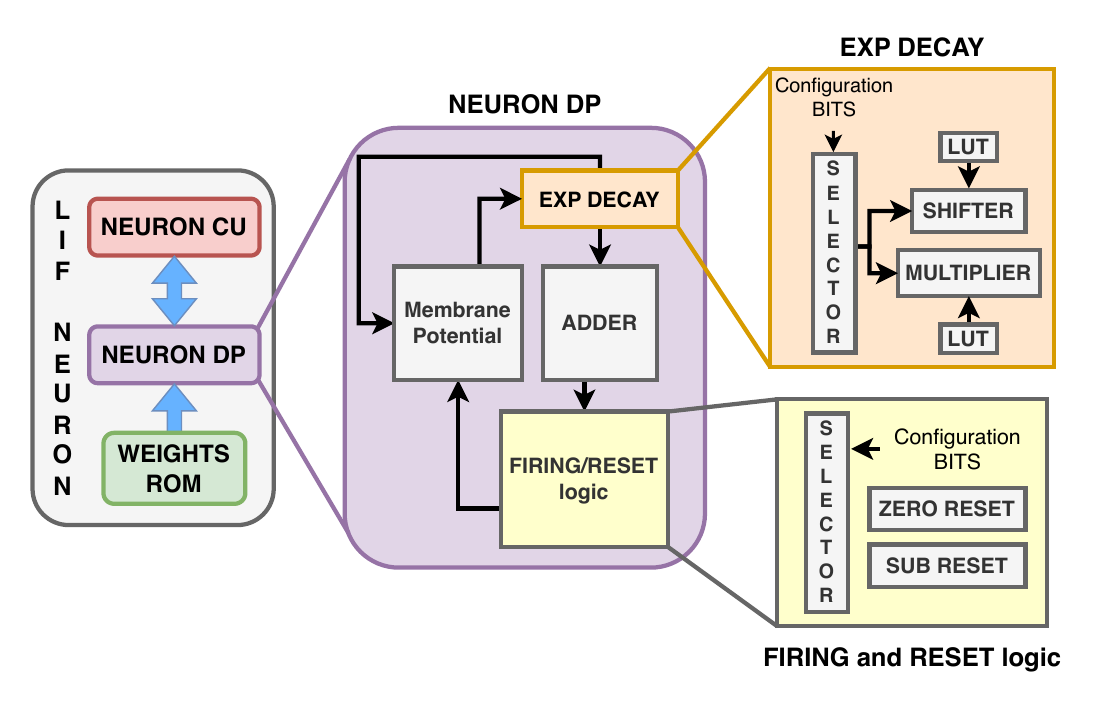}}
\caption{High-level schematic of the hardware implementation of the LIF neuron. The top entity (left) includes a Control Unit (CU) and a Data Path (DP), along with a weights ROM for parameter storage. Within the DP (center), the membrane potential is updated by an exponential decay stage and an adder, and the neuron’s firing/reset logic is managed by configurable blocks. The exponential decay can be computed via either a shifter or a multiplier (right, orange box), accessed from different LUT entries according to user-defined bits, while the firing/reset logic (right, yellow box) also adapts to configurable bits that select between different reset strategies (e.g., zero or subtract reset).
%High-level schematic of the LIF neuron hardware. The top entity includes a Control Unit (CU), a Data Path (DP), and a weights ROM. The DP performs exponential decay and reset operations, with the use of a shifter or multiplier and zero or subtract reset selected via configuration bits. All options are set before synthesis, ensuring design flexibility without the overhead of runtime reconfiguration.
}
\label{fig:neuron_structure}
\end{figure}

The hardware implementation of the different neurons is designed with modularity in mind, comprising a top entity, a \gls{DP} and a \gls{CU}. Rather than analyzing a complete system architecture, our focus is on a fine-grained study of the architecture of a single neuron. In this design, the top entity handles communication with memory, where parameters are stored and receives spikes from external sources. The \gls{DP} processes the data, including updating the membrane potential and checking firing conditions, based on control signals generated by the \gls{CU}. 
This modular approach was designed and implemented in SystemVerilog, targeting Xilinx FPGAs, and a high-level representation of the neuron structure is shown in Figure~\ref{fig:neuron_structure}.
Moreover in our implementation, no \gls{DSP} or \gls{BRAM} blocks were used. Although modern \gls{FPGA}s provide \gls{DSP} slices, their use is not guaranteed by the synthesis tool and can vary based on tool optimizations, synthesis directives, and resource availability. To ensure a fair and generalizable comparison, we enforced the use of \gls{LUT}-based arithmetic across all neuron variants. This avoids backend-specific optimizations and enables consistent post-synthesis evaluation of area and power, independent of device-specific heuristics.

\subsection{Clock-Driven Implementation}

In the clock-driven implementation, inputs are received as a string of bits, where a '0' indicates no spike and a '1' indicates a spike. During each time step, if all inputs are zero, it means no input spikes are present, and the neuron simply updates its membrane potential. When a time step contains at least one input different from zero, the neuron processes the inputs serially, executing the mathematical calculations described in Section \ref{sec:math_model_neuron}. While pipelining the serial datapath could increase throughput by overlapping operations within a single time step, we chose a non-pipelined design to maintain consistency across all six neuron variants. This ensures a fair comparison of architectural trade-offs without introducing implementation-dependent optimizations that would affect area and power. The goal of this work is not to maximize absolute performance, but to isolate and compare the inherent efficiency of each neuron processing strategy under equivalent design conditions.

\subsection{Event-Driven Implementation}

In the serial case, when a time step consists entirely of zeros (indicating no input spikes), as for the clock-driven neuron, the neuron increments a counter to keep track of the time since the last spike. On the other hand, when the neuron is configured to operate with \gls{AER} data, the system uses the information embedded into each \gls{AER} string to access the memory, and a register is used to keep track of the time.

\section{Simulation and Results Analysis}\label{sec:results}

We performed simulations to evaluate performance metrics for both clock-driven and event-driven neuron models varying the level of sparsity. Data were generated in Serial and \gls{AER} formats, with sparsity varying along the two dimensions: Input Sparsity and Temporal Sparsity. We focused on processing mode, sparsity, and decay implementation (shifter vs. multiplier) because these are among the most influential factors in hardware performance. This choice enables a targeted evaluation of trade-offs in latency, energy, and area under realistic constraints. %The simulation run for 100 time steps, covering temporal sparsity levels from 5\% to 100\% in increments of 5\%, using 8 input channels. 

All designs run at 100 MHz, to align with data reported in previous publications on \gls{FPGA}-specific hardware accelerators such as \cite{gupta_fpga_2020} or \cite{liu_low_2023}. Because power and latency scale almost linearly with frequency, changing the clock would shift absolute values but not the relative trends across our six neuron variants. We simulate 100 time-steps and drive each neuron with 8 input channels. Latency, power and resource usage grow linearly with extra steps or channels, so varying them would not change the qualitative conclusions. A testbench capable of interfacing with both the clock-driven and event-driven neuron models was developed, allowing a direct comparison of their performance.
All neurons use 6-bit weights and biases, a common quantization level in low-power designs that balances accuracy and hardware efficiency without the complexity of more aggressive schemes (e.g., 4 bits). The membrane potential is stored in 9 bits to accommodate the accumulation of multiple weighted inputs, which is particularly important in event-driven operation. A 7-bit counter encodes simulation time (up to 128 steps), while a 3-bit input address field supports the 8 input channels used in our experiments.

\subsection{Analysis Process}

The system simulation and performance analysis consist of three stages: Functional Simulation evaluates latency across varying temporal sparsity levels to assess neuron processing time; Post-Synthesis Simulation gathers switching activity data - using \gls{SAIF} -  from synthesized architectures for power consumption analysis; Implementation Analysis examines hardware implementation to measure power consumption and resource utilization under different conditions.

\subsection{Analysis Process - Latency}\label{subsec:analysis_process_latency}

\begin{figure*}[t]
\centerline{\includegraphics[width=\textwidth]{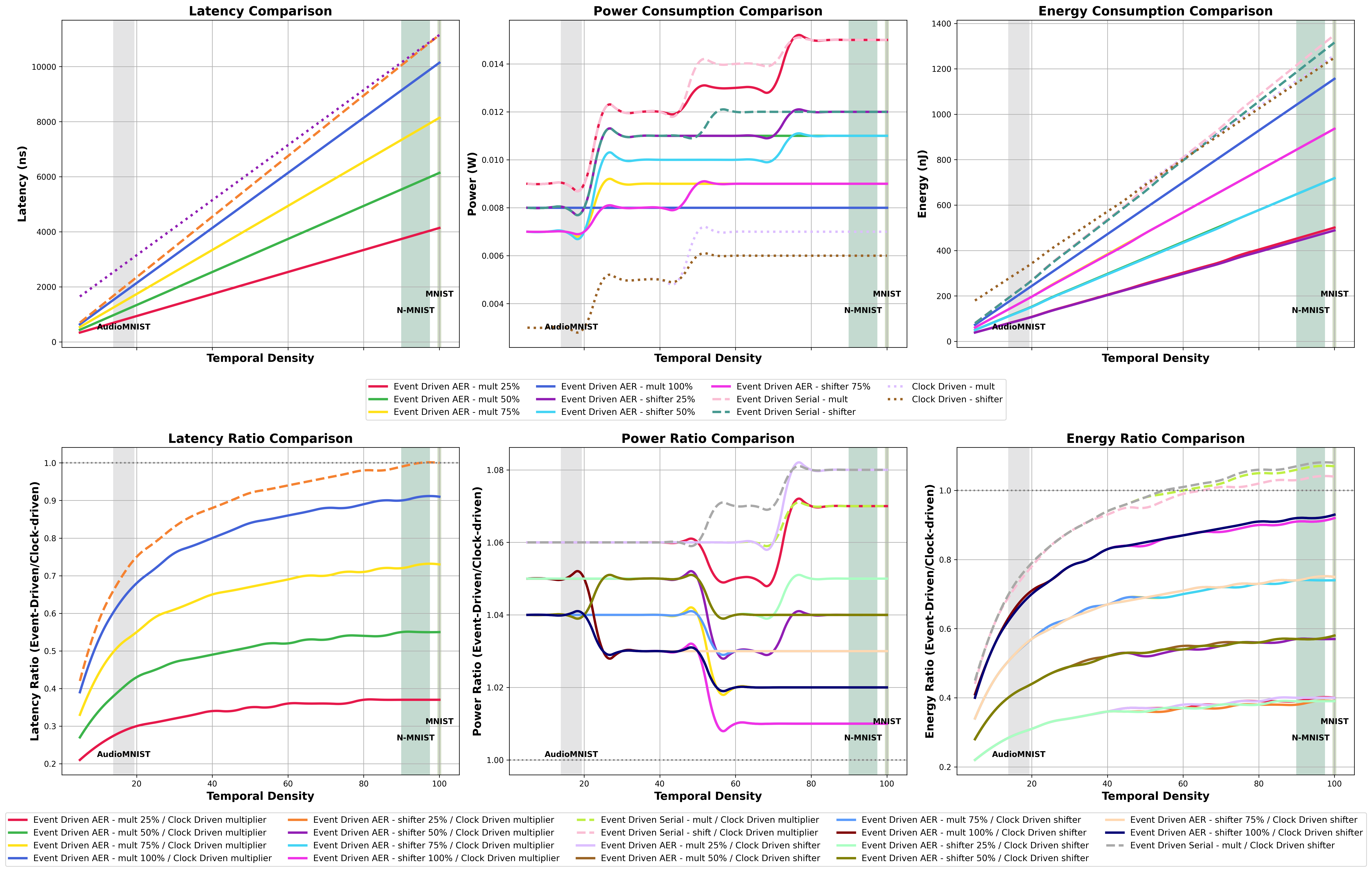}}
\caption{Comparison of latency, power consumption, and energy across six neuron architectures under varying temporal input density. The top row shows absolute values for all neuron configurations, while the bottom row shows the normalized ratio of event-driven to clock-driven performance (multiplier-based as baseline). Vertical shaded bands mark the average temporal densities of the three evaluated datasets: AudioMNIST (low), N-MNIST (moderate), and MNIST (high). The Figure illustrates the key trade-offs between design choices under realistic sparsity scenarios, emphasizing that energy and latency benefits are most pronounced under sparse input conditions, while power advantages may depend on specific implementation details such as control overhead.
}
\label{fig:all_metrics}
\end{figure*}

To evaluate the impact of temporal sparsity on computational performance, simulations measured the total processing time while varying the sparsity levels and recording the overall latency for each configuration.
In our experiments, architectures that process data serially showed constant latency across the sparsity sweep, because they process every time-step regardless of how many inputs are active. In contrast, the \gls{AER} neuron was directly influenced by input sparsity, so to capture this effect, simulations were conducted at different levels (25\%, 50\%, 75\%, and 100\%). Both decay evaluation methods configurations yielded identical latency performance. Therefore, we focused our analysis on the multiplier-based architecture.
%, noting that results would be identical using the shifter mechanism. %Lastly, all the simulations were performed with a 100 MHz Clock to align with data reported in previous publications on \gls{FPGA}-specific hardware accelerators such as \cite{gupta_fpga_2020} or \cite{liu_low_2023}.

The left-hand side of Figure~\ref{fig:all_metrics} presents two related views. The top panel shows the absolute latency curves: the clock-driven model consistently exhibits the highest latency, especially under low time sparsity conditions since it processes all time steps without optimization. In contrast, the \gls{AER} neuron, which selectively processes only active inputs, achieves markedly lower latency when Inputs are more sparse. However, as input density increases, the performance advantage of the event-driven approach diminishes.

The bottom panel unifies the analysis by plotting the ratio \( R_{i,k}(\text{Sparsity}) = \frac{L_{\mathrm{event}_i}}{L_{\mathrm{clock}_k}} \) where \textit{i} denotes different event-driven configurations and \textit{k} represents the two clock-driven configurations (multiplier/shifter-based). This formulation explicitly highlights the comparison across multiple configurations for both paradigms. This ratio clearly illustrates the effect of input sparsity: when inputs are mostly inactive, $R_{i,k}(\text{Sparsity})$ is well below 1, highlighting the efficiency of event-driven processing. On the other hand, as input density increases, the ratio approaches or exceeds 1, indicating that the overhead of managing frequent events erodes the latency benefit.

\subsection{Analysis Process - Power}

After conducting the performance analysis of the simulated architecture, the design was synthesized, and post-synthesis simulations were carried out to extract the \gls{SAIF} files. The target platform for implementation was the Pynq-Z2 \gls{FPGA}, featuring an ARTIX-7 equivalent and a 650 MHz dual-core Cortex-A9 processor. The synthesis effort adhered to Vivado's standard level specifications. Power consumption measurements were performed using the generated \gls{SAIF} files under stimulus conditions identical to those in the simulation phase. This analysis was conducted on both implementations, as the different hardware solutions impact the power consumption metrics.

Vivado Power Analysis considers two primary contributions to total power consumption: Static and Dynamic Power. However, while Vivado reports a static power of 106 mW across all configurations, this consistency is expected. Static power on the Artix-7 FPGA is determined by device-level characteristics such as leakage current and clocking infrastructure, which remain unchanged given our small design footprint. All neuron variants occupy only a fraction of the fabric, so variations in logic utilization (e.g., between shifter and multiplier designs) do not meaningfully affect static power estimates. As a result, we excluded static power from our analysis to focus on dynamic contributions, which more accurately reflect architectural differences.
%Vivado reports a constant static power of 106mW across all configurations, which is expected given the small design size. On Artix-7 FPGAs, static power mainly depends on device-level factors like leakage and clocking, not minor logic variations. We therefore focus on dynamic power, which better captures architectural differences.

Most of the observed results align with intuitive expectations (e.g., higher input activity leads to increased resource usage). Yet, one finding stands out as unpredictable: in the \gls{AER}-based architecture, higher Input Sparsity is correlated with increased power consumption. Initially, one might assume that fewer input spikes should reduce overall activity—and thus lower power—but our measurements show the opposite. A plausible explanation lies in the overhead of the control circuitry. In the serial case, the CU processes a fixed-width input vector at each time step, resulting in predictable control activity that scales linearly with the number of channels. In contrast, in the AER case, the CU remains idle until a packet is detected; when this happens, it triggers a series of actions: decoding the address, initiating memory access, and updating the membrane potential causing a burst of control activity, including FSM transitions and memory interactions. Even though the neuron \gls{DP} itself processes fewer spikes, the event-based state transitions in the control logic appear to overshadow any power savings from the neuron’s reduced arithmetic operations. In other words, the power cost of these “control events” dominates at high input sparsity, producing a counterintuitive increase in total power consumption.
Both serial event-driven architectures (multiplier-based and shifter-based) exhibit a trend of rising power consumption as temporal sparsity decreases, which is consistent with a design whose activity scales linearly with the number of input events. Notably, the second column of the first row in Figure~\ref{fig:all_metrics} clearly demonstrates that the shifter-based design significantly reduces power consumption compared to the multiplier-based approach.%, underscoring the advantages of using bit-shift operations to approximate the decay factor.

As expected for serial, clock-driven designs, dynamic power rises as input sparsity decreases, even though their latency remains flat (see Section \ref{subsec:analysis_process_latency}), because more input spikes trigger additional arithmetic activity in each cycle. This increase is due to the more complex logic, including input accumulation and firing conditions, becoming active more frequently as more inputs generate events, contrasting with the simpler decay logic that predominates under high sparsity. While the results reveal no significant difference between the shifter-based and multiplier-based mechanisms for the clock-driven architectures, this is largely due to the precision used by Vivado (0.001W). However, even with this small difference, the power gap becomes more pronounced in a full network implementation, where the impact of power optimization strategies on the entire system is amplified.

\subsection{Analysis Process - Energy Analysis}

The last measured metric is energy; it is defined as the product of power and latency ($E = P \times L$). Energy provides a comprehensive quantitative measure of the computational requirements to process inputs in a neuron beyond a simple comparison of power or latency individually.

Figure~\ref{fig:all_metrics} (column 3, row 1) shows the absolute energy consumption for each configuration. The \gls{AER}-based architectures exhibit an increasing trend in energy as the number of active inputs grows, which reflects the heightened activity in the \gls{AER} control components under denser input conditions. In contrast, the clock-driven and serial architectures follow a similar linear pattern but maintain higher overall energy values under low-sparsity conditions.%, as both their power consumption and latency remain elevated relative to the event-driven counterparts.

%This analysis confirms our previous findings: under highly sparse conditions, event-driven designs offer a significant energy advantage, whereas this benefit diminishes as the density of inputs increases.

\subsection{Resources Utilization analysis}

\begin{figure}[t]
\centerline{\includegraphics[width=\columnwidth]{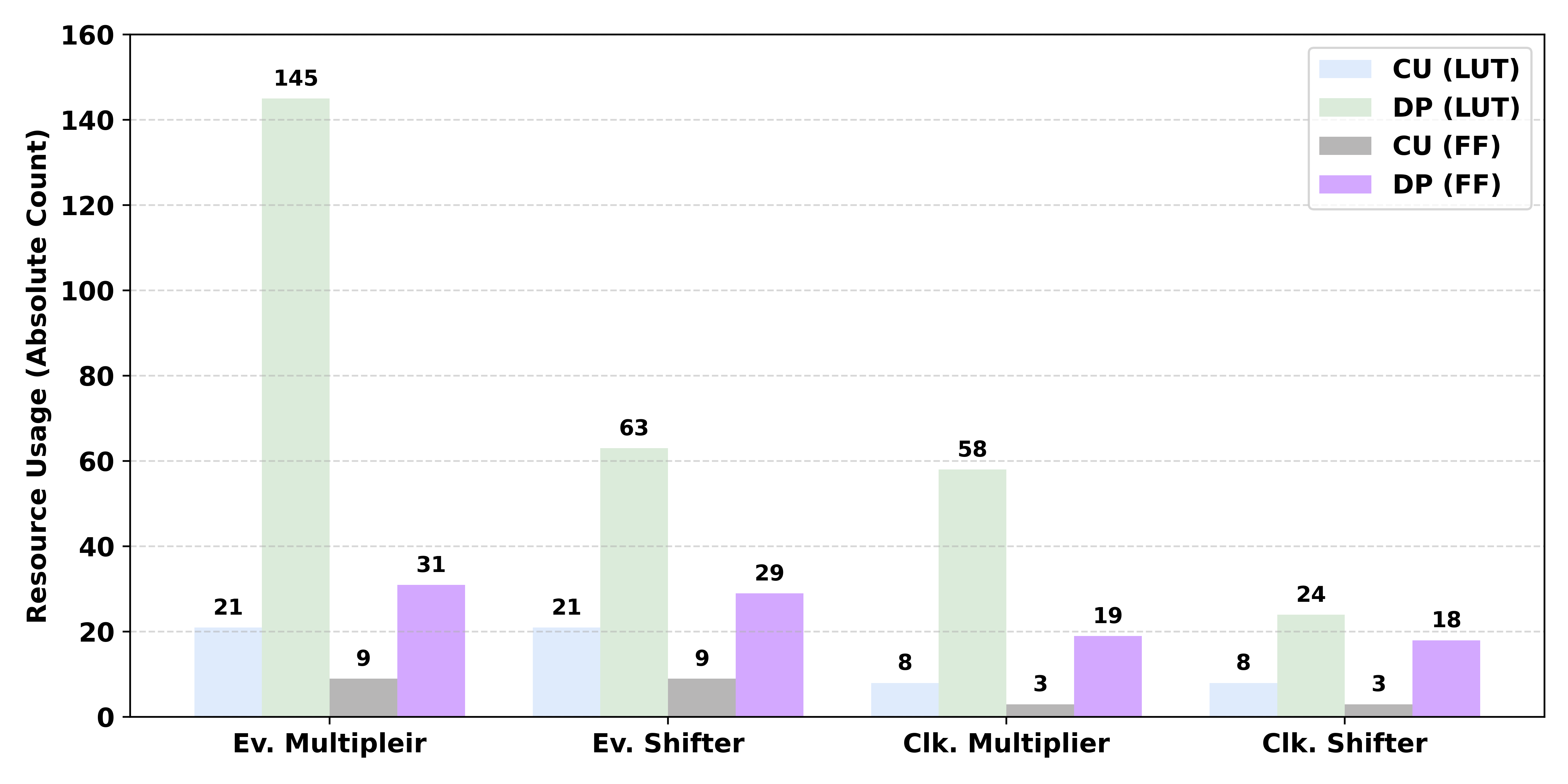}}
\caption{Resource utilization comparison for different neuron implementations, including event-driven multiplier, event-driven shifter, clock-driven multiplier, and clock-driven shifter. The figure presents the total number of utilized resources, specifically LUTs, and FFs, while also distinguishing their respective contributions to the overall resource consumption.}
\label{fig:resources_utilization}
\end{figure}

Figure~\ref{fig:resources_utilization} illustrates the distribution of \glspl{FF} and \glspl{LUT} across the \gls{CU} and \gls{DP} for the four designs. In the case of an event-driven neuron, the choice between communication methods (i.e., \gls{AER} versus serial) does not significantly affect resource usage apart from a difference in the I/O pins of the architecture. For this reason, it was not considered in the comparison. The event-driven architecture using a multiplier exhibits the highest overall resource utilization. %, owing to the intrinsic complexity of its multiplier logic. 
In contrast, the event-driven shifter implementation employs significantly fewer \glspl{LUT} in the \gls{DP}, while its \gls{CU} usage remains unchanged compared to the multiplier-based version. Similarly, the clock-driven designs in both configurations show a simpler footprint than the event-driven shifter implementation, and follow the same trend observed between the multiplier- and shifter-based event-driven architectures.

\section{Conclusion and Future Works}\label{sec:future_works}

This paper compared multiplier-based and shifter-based event-driven LIF neurons against clock-driven implementations, evaluating software and hardware performance across resource usage, latency, power, and accuracy. Although no implementation clearly outperformed the others, results highlight the importance of balancing resource constraints, accuracy, and operating conditions. Future work could integrate these neurons into dedicated hardware accelerators, explore quantization effects, and analyze more complex neuron models to further clarify optimal configurations.

%\bibliographystyle{IEEEtran}
%\bibliography{bibliography}
\printbibliography

\end{document}